\documentclass[pmlr]{jmlr}
\RequirePackage{graphicx}
\usepackage{amsmath}
\usepackage{amssymb}
\usepackage{mathrsfs}
\usepackage{hyperref}
\usepackage{booktabs}
\usepackage{multirow}
\usepackage{longtable}
\makeatletter
\def\set@curr@file#1{\def\@curr@file{#1}} 
\makeatother
\usepackage[load-configurations=version-1]{siunitx} 

\theorembodyfont{\upshape}
\theoremheaderfont{\scshape}
\theorempostheader{:}
\theoremsep{\newline}

\jmlrvolume{298}
\jmlryear{2025}
\jmlrworkshop{Machine Learning for Healthcare}

\title[Multimodal Learning for Early Detection of Pancreatic Cancer]{Early Detection of Pancreatic Cancer Using Multimodal Learning on Electronic Health Records}

\author{\Name{Mosbah Aouad}
       \Email{maouad2@illinois.edu}\\ 
       \addr University of Illinois Urbana-Champaign
       \AND
       \Name{Anirudh Choudhary}
       \Email{ac67@illinois.edu}\\ 
       \addr University of Illinois Urbana-Champaign
       \AND
       \Name{Awais Farooq}
       \Email{afaroo32@uic.edu}\\ 
       \addr University of Illinois Chicago
       \AND
       \Name{Steven Nevers}
       \Email{snevers@uic.edu}\\ 
       \addr University of Illinois Chicago
       \AND
       \Name{Lusine Demirkhanyan}
       \Email{lusinhd@uic.edu}\\ 
       \addr University of Illinois Chicago
       \AND
       \Name{Bhrandon Harris}
       \Email{bharri35@uic.edu}\\ 
       \addr University of Illinois Chicago
       \AND
       \Name{Suguna Pappu}
       \Email{spappu@illinois.edu}\\ 
       \addr University of Illinois at Urbana-Champaign
       \AND
       \Name{Christopher Gondi}
       \Email{gondi@uic.edu}\\ 
       \addr University of Illinois Chicago
       \AND
       \Name{Ravishankar Iyer}
       \Email{rkiyer@illinois.edu}\\ 
       \addr University of Illinois Urbana-Champaign}

\begin{document}

\maketitle

\begin{abstract}
  Pancreatic ductal adenocarcinoma (PDAC) is one of the deadliest cancers, and early detection remains a major clinical challenge due to the absence of specific symptoms and reliable biomarkers. In this work, we propose a new multimodal approach that integrates longitudinal diagnosis code histories and routinely collected laboratory measurements from electronic health records to detect PDAC up to one year prior to clinical diagnosis. Our method combines neural controlled differential equations to model irregular lab time series, pretrained language models and recurrent networks to learn diagnosis code trajectory representations, and cross-attention mechanisms to capture interactions between the two modalities. We develop and evaluate our approach on a real-world dataset of nearly 4,700 patients and achieve significant improvements in AUC ranging from 6.5\% to 15.5\% over state-of-the-art methods. Furthermore, our model identifies diagnosis codes and laboratory panels associated with elevated PDAC risk, including both established and new biomarkers. Our code is available at \url{https://github.com/MosbahAouad/EarlyPDAC-MML}. 
  
\end{abstract}

\section{Introduction}

Pancreatic ductal adenocarcinoma (PDAC) is one of the deadliest cancers, with a five-year survival rate of approximately 10\% (\citealp{leiphrakpam2025trends}). Despite its relatively low incidence, PDAC is expected to become the second leading cause of cancer-related deaths in the U.S. by 2030 (\citealp{espona2024overcoming}). Most patients are diagnosed at advanced or metastatic stages because symptoms do not manifest in the early stages (\citealp{singhi2019early}). Existing screening protocols are limited to high-risk individuals, excluding a substantial portion of the affected population (\cite{blackford2024pancreatic}). 

The increasing availability of longitudinal electronic health records (EHRs) provides an opportunity to improve PDAC detection in broader populations. Our goal is to develop a multimodal approach that integrates sequential laboratory measurements and disease code histories for early PDAC detection.  We address this goal by modeling laboratory trajectories as irregular time series using neural differential equations (\cite{kidger2020neural}), encoding diagnosis codes sequence with a bidirectional recurrent model, and apply a cross-attention fusion mechanism to capture complementary information across modalities.  

Prior studies demonstrated that machine learning (ML) can identify predictive patterns for early detection of PDAC from EHRs (\citealp{jia2023pancreatic}). However, existing approaches rely on a single modality, either using diagnostic codes (\citealp{appelbaum2021development, placido2023deep}) or laboratory tests (\citealp{haab2024rigorous, cichosz2024prediction}), failing to capture the cross-modal interactions.  Labs provide high-resolution, continuous signals that reveal early physiological changes such as rising glucose levels  months before PDAC symptoms appear (\citealp{sharma2018fasting}). In contrast, diagnostic codes capture clinically validated events, including comorbidities, symptoms, and imaging findings, that labs may miss. Thus, integrating these modalities provides complementary information on patient history, resulting in a comprehensive multimodal representation of disease progression.

However, simple fusion of lab measurements and diagnosis codes (e.g, concatenation) presents two key challenges. First, the modalities differ in temporal resolution and structure: lab tests constitute continuous irregular time series, while diagnosis codes capture sparse, discrete clinical events. This mismatch necessitates modality-specific temporal encoding prior to fusion. Second, frequently ordered lab tests are collected at varying frequencies and reflect distinct physiological systems, such as hepatic and metabolic functions, that evolve on different timescales over the disease trajectory (\citealp{janssens2013managing}). Consequently, the model must learn time-varying lab representations while incorporating mechanism to capture cross-modal dependencies.  

To address these challenges, we introduce a new multimodal approach for early PDAC detection. Firstly, we group lab measurement trajectories into panels and model each panel as multivariate irregular time series. Each panel comprises related lab measurements reflecting a common physiological system, allowing the model to learn system-specific dynamics. We encode each panel trajectory using neural controlled differential equations (NCDEs) to capture disease progression while handling irregular sampling times and missing data. Cross-panel dependencies are modeled via a self-attention mechanism.  Secondly, we generate context-specific embeddings for diagnosis codes using pretrained language models and model their temporal evolution with a bidirectional Long Short-Term Memory (Bi-LSTM). Lastly, we employ cross-attention mechanism to integrate diagnostic and laboratory representations into a unified, temporally-aligned cross-modal representation for pancreatic cancer risk prediction. We evaluate the proposed approach against state-of-the-art PDAC detection models on an EHR data from OSF Saint Francis Medical Center, comprising $4,700$ patients with up to $14$ years of history. Our method significantly outperforms baselines on early PDAC detection across multiple time horizons, achieving AUC improvements of up to $12.33\%$ , $6.5\%$ and $15.5\%$ for six, nine, and twelve-month prior to PDAC diagnosis, respectively. Moreover, the model highlights relevant high-risk biomarkers, including acute and chronic pancreatitis diagnoses, and changes in hepatic and metabolic lab panels, validating its potential in early PDAC screening.

In summary, our contributions are:
\begin{enumerate}  
 \item  A novel multimodal approach for early PDAC detection that leverages cross-attention to capture interactions between longitudinal diagnosis code trajectories and routine laboratory measurements. 
    \item A panel-specific laboratory measurement representation approach that captures physiological system-level dynamics as well as inter-panel dependencies using self attention.
    \item We demonstrate substantial AUC improvements of upto 12\%, 6\%, and 15\% over 6, 9, and 12 months time horizon for PDAC detection compared to state-of-the-art models, while also validating clinically relevant biomarkers.
\end{enumerate}

\subsection*{Generalizable Insights about Machine Learning in the Context of Healthcare}

Our work provides insights into modeling multimodal, irregularly sampled clinical time series for healthcare applications. We show that grouping laboratory measurements into clinically meaningful panels improves representation learning by capturing function-specific temporal dynamics. Furthermore, our architecture demonstrates that combining modality-specific encoders with cross-attention mechanisms enables the model to capture latent dependencies across modalities. Our findings extend beyond PDAC detection, offering a framework for leveraging multimodal EHR data in predictive modeling for other complex diseases

\section{Related Work}

\paragraph{Machine Learning for Early Detection of PDAC:}There has been growing interest in applying ML methods to different types of EHR data for pancreatic cancer risk assessment at scale in recent years. Early studies leveraged the occurrence of specific diagnosis codes in a large EHR database and developed shallow ML models, including Extreme Gradient Boosting (\citealp{chen2023novel,chen2021clinical}) and regularized logistic regression (\citealp{appelbaum2021development}), to infer cancer risk. However, these approaches did not account for the temporal dynamics in patients' longitudinal histories. To address that, \citealp{placido2023deep} represented patients' disease trajectories as a longitudinal sequence of clinical events and introduced a transformer architecture with an updated positional encoding, improving predictive performance for cancer risk recurrence. While these diagnosis code-based methods capture broad disease trends, they overlook fine-grained changes in patients' histories, which are better reflected in laboratory measurements. To address this limitation, \citealp{park2023structured} proposed a group-based neural network in which labs are grouped based on related organ function, a dedicated neural network processes the inputs of each group, and a final layer concatenates the feature representations of the different groups and projects them into a shared space for PDAC risk prediction six months in advance. However, this approach achieved limited performance due to its reliance on cross-sectional lab measurements at a single encounter, missing crucial longitudinal trends in patient data. \citealp{park2022deep} introduced a group-based neural network in which individual small networks model the longitudinal measurements of each lab as irregular time series data, employing a Poisson-based masking algorithm to account for random missingness in measurements, thus achieving significant improvements in early PDAC prediction. Despite those advancements, no prior work has proposed a multimodal approach that integrates both clinical trajectories from diagnosis codes and longitudinal laboratory measurements for the early detection of PDAC. Our work addresses this gap by combining these two complementary data sources, leveraging their interactions to improve predictive performance. 
\paragraph{Modeling of Irregular Health Time Series:} Several methods have been proposed to model irregular time series within healthcare applications. GRU-D incorporates exponential decay in hidden states to capture temporal patterns (\citealp{che2018recurrent}). SeFT models each observation individually using a set function approach and then aggregates their representations(\citealp{horn2020set}). RAINDROP represents data as distinct sensor graphs to handle irregular temporal dependencies (\citealp{zhang2021graph}). mTAND utilizes a multi-time attention mechanism to learn from irregularly sampled data (\citealp{shukla2021multi}). For continuous-time modeling, neural ODEs employ neural networks to approximate complex differential equations, enabling effective interpolation and extrapolation (\citealp{chen2018neural}). ODE-RNN expands on neural ODEs by incorporating new observations to update the hidden states of an RNN (\citealp{rubanova2019latent}). Neural CDEs extend the concept of neural ODEs by incorporating control paths to model irregular time series data (\citealp{kidger2020neural}). This approach has demonstrated state-of-the-art performance in handling irregular time series functions compared to the other methods.  In our work, we leverage Neural CDEs within a new clinically informed model architecture that focuses on grouping lab measurements into functional panels and learning interactions between panels. 

\section{Methods}

Our approach is motivated by four key observations: 1) Pancreatic cancer often remains asymptomatic until advanced stages, resulting in sparse and irregular clinical measurements for majority of patients. 2) Patients engage with the healthcare system at varying frequencies based on disease severity, comorbidities and individual behaviour, resulting in variable-length clinical histories. 3) Routine lab tests are ordered in panels that reflect distinct physiological systems (e.g., hepatic or metabolic). Modeling these panels separately aligns with diagnostic workflows and enables learning function-specific patterns. 4) Capturing interactions between lab trajectories and diagnostic codes would allow the model to integrate complementary signals - labs often indicate early physiological changes during cancer progression, while diagnoses reflect  clinically significant events.

\subsection{Problem Setup}

For the early detection of PDAC, we consider a dataset $\mathcal{D} = \{(X_i, Y_i)\}_{i=1}^{N}$, comprising $N$ patients. For each patient $i$, $X_i \in \mathcal{X}$ denotes the patient's longitudinal record, and $Y_i \in \{0,1\} \subseteq \mathcal{Y}$ is a binary outcome indicating whether the patient was diagnosed with PDAC within a certain time following the last recorded observation in $X_i$. Each longitudinal record is defined as $X_i = (C_i, L_i)$, where $C_i = [c_{i,1}, c_{i,2}, \ldots, c_{i,m}]$ is a sequence of $m$ diagnosis code observations, and $L_i = [l_{i,1}, l_{i,2}, \ldots, l_{i,k}]$ is a sequence of $k$ laboratory test observations. The lengths $m$ and $k$ vary across patients depending on the duration and frequency of healthcare utilization. Each diagnosis observation $c_{i,j}$ is represented as a tuple $c_{i,j} = (t_{i,j}, d_{i,j})$, where $t_{i,j}$ is the timestamp of the event and $d_{i,j}$ is a diagnosis code assigned according to International Classification of Diseases, 10th Revision (ICD-10) standards (\citealp{WHO2019ICD10}). Each lab observation $l_{i,j}$ is encoded as a quadruple $l_{i,j} = (t_{i,j}, o_{i,j}, s_{i,j}, f_{i,j})$, where $t_{i,j}$ is the time of encounter, $o_{i,j} \in \mathcal{O}^v$ is a vector of $v$ lab tests conducted at that encounter, $s_{i,j} \in \mathcal{S}^v$ contains the corresponding lab measurements, and $f_{i,j} \in \mathcal{F}^v$ represents the number of recorded instances of each lab up to time $t_{i,j}$. Our primary objective is to develop a multimodal prediction model $\phi: \mathcal{X} \rightarrow \mathcal{Y}$ that accurately identifies patients at elevated risk of PDAC prior to clinical diagnosis, while also validating informative labs and diagnosis code-based biomarkers.

\begin{figure}[t]
\includegraphics[width=\textwidth]{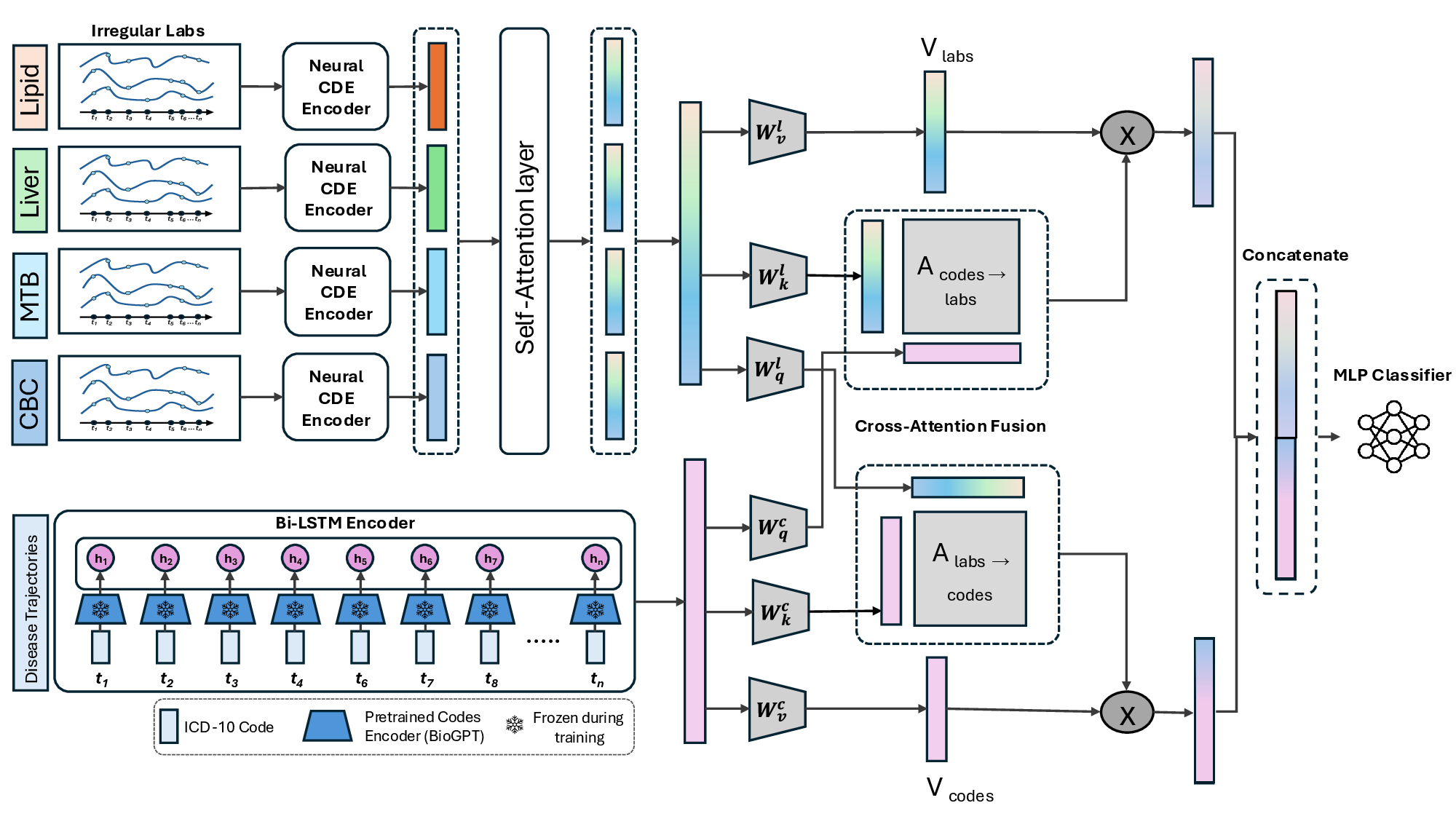}
\caption{\textbf{Proposed Network Architecture}: The model comprises three main modules: (1) a labs-feature extractor that groups labs into four panels (metabolic, CBC, lipid, and livers), generates panel-based feature vectors using NCDE encoders, and models inter-panel interactions via a self-attention network; (2) a diagnosis-codes feature extractor that encodes chronologically reported diagnosis codes into a feature representation of disease trajectories; and (3) a cross-attention mechanism that captures interactions between labs and codes to produce a final feature vector, which is then passed to a multilayer-perceptron (MLP) classifier for early detection of PDAC.}
\label{fig:network-architecture}
\end{figure}

\subsection{Model Architecture}

Figure~\ref{fig:network-architecture} provides an overview of the model architecture, which comprises three core components: a labs-feature extractor, a diagnosis-codes-feature extractor , and a cross-attention fusion module. Each component is described in detail below.

\subsubsection{Labs-Feature Encoder}

This module is designed to extract meaningful patterns from each patient's irregularly sampled laboratory time series. Each patient trajectory $L_i = [l_{i,1}, l_{i,2}, \ldots, l_{i,k}]$ contains measurements of up to 35 routinely collected laboratory tests, which are grouped into four standard clinical panels: metabolic, Complete Blood Count (CBC), lipid, and liver. These panels are commonly used to assess different physiological functions. To better capture function-specific patterns and reduce inter-panel redundancy across related labs, we organize lab measurements into four clinical panels with distinct sub-trajectories: $L_i^p = [l_{i,1}^p, l_{i,2}^p, \ldots, l_{i,k_p}^p]$, where $p \in \{\textit{metabolic}, \textit{CBC}, \textit{liver}, \textit{lipid}\}$ and $k_p$ represents panel-specific observation count. Each $L_i^p$ contains time-stamped observations of tests within the corresponding panel.

We model each panel sub-trajectory using controlled differential equation to capture the continuous-time evolution of patient physiology while handling irregularly sampled time series. Specifically, we leverage NCDE networks (\citealp{kidger2020neural}) to encode each lab panel $L_i^p$. We first transform the sequence of lab observations into a continuous path, $\mathcal{L}^p:[t_1,t_k] \rightarrow \mathbb{R}^{v+1}$ using natural cubic spline with interpolation points $\mathcal{L}^p_{t_j}=(l_j^p, t_j)$ for an observation at time $t_j$. NCDE utilizes a neural network \( \phi_\theta : \mathbb{R}^w \to \mathbb{R}^{d \times (v+1)} \) to parameterize the time-varying dynamics of hidden state $z_t^p$, where \( d \) is the hidden feature dimension. The hidden state evolves according to the following integral defined on controlled differential equation:
\[
z_t^p = z^p_{t_0} + \int_{t_0}^{t} \phi_\theta(z^p_s) \, d\mathcal{L}_s^p, \quad \text{for } t \in (t_0, t_n],
\]
with the initial hidden state given by $z^p_{t_0} = \zeta_\theta(L^p_0, t_0)$, where \( \zeta_\theta : \mathbb{R}^{v+1} \to \mathbb{R}^d \) is another neural network that applies a non-linear transformation to the initial condition to ensure translational invariance. The final hidden state $z_{t_n}^p$ summarizes the entire panel trajectory and serves as its learned feature representation. 
To enhance the model's ability to capture clinical decision patterns, we incorporate observational intensity as an additional input channel (\citealp{morrill2021neural}) to the model. This intensity reflects the frequency of clinical measurements under the assumption that higher frequency values may indicate greater patient risk. By adding observational intensity, the model can learn to associate irregular observation patterns with underlying patient behavior and clinical outcomes.

To capture dependencies across physiological systems, we treat the four panel-specific embeddings as input tokens to a self-attention mechanism. Let \( \{z^p\}_{p=1}^4 \), with \( z^p \in \mathbb{R}^d \), denote the set of panel-wise representations. These are stacked into a matrix \( Z \in \mathbb{R}^{4 \times d} \), which serves as the input to a self-attention layer:

\begin{equation}
    \tilde{Z} = \text{SelfAttention}(Z) = \text{softmax}\left(\frac{QK^\top}{\sqrt{d}}\right)V
\end{equation}

where \( Q = ZW_Q \), \( K = ZW_K \), \( V = ZW_V \), and \( W_Q, W_K, W_V \in \mathbb{R}^{d \times d} \) are learnable projection matrices. The output \( \tilde{Z} \in \mathbb{R}^{4 \times d} \) contains updated panel embeddings with contextual information from other panels.
The attention mechanism enables the model to capture both shared and distinct dynamics across physiological systems and refine each panel embedding accordingly. To obtain a unified lab representation, we apply mean pooling over the context-aware embeddings:
\begin{equation}
    z_i^{\text{labs}} = \text{MeanPool}(\tilde{Z}) \in \mathbb{R}^d
\end{equation}

The resulting vector \( z_i^{\text{labs}} \) serves as the patient's lab representation and is used in the subsequent cross-modal fusion with diagnostic code features.

\subsubsection{Diagnosis-Code Feature Encoder}
To model the diagnostic history of each patient \( i \), we represent their diagnosis sequence as \( C_i = [c_{i,1}, c_{i,2}, \ldots, c_{i,m_i}] \), where each clinical event \( c_{i,j} = (t_{i,j}, d_{i,j}) \) includes a timestamp \( t_{i,j} \) and a diagnosis code \( d_{i,j} \), assigned using the ICD-10 standard.

Each diagnosis code \( d_{i,j} \) is mapped to a semantic embedding using pre-trained BioGPT \citep{luo2022biogpt}, a biomedical language model trained on PubMed abstracts and full-text articles. BioGPT embeddings capture contextual relationships between codes grounded in the biomedical literature, allowing the model to incorporate richer medical semantics than one-hot or static embeddings. Since BioGPT produces high-dimensional outputs, we apply a trained autoencoder (\citealp{kane2023compressed}) to reduce each embedding to a lower-dimensional space \( e_{i,j} \in \mathbb{R}^h \), preserving semantic information while reducing computational complexity.

The sequence of compressed embeddings \( E_i = [e_{i,1}, e_{i,2}, \ldots, e_{i,m_i}] \) is then passed through a bidirectional Long Short-Term Memory (Bi-LSTM) network (\cite{huang2015bidirectional}). Bi-LSTM is well-suited for modeling diagnosis sequences because it captures temporal dependencies in both forward and backward directions, which is important in clinical settings where diagnoses accumulate over time and past conditions may influence future ones. The Bi-LSTM outputs a sequence of hidden states \( H_i = [h_{i,1}, h_{i,2}, \ldots, h_{i,m_i}] \), where each \( h_{i,j} \in \mathbb{R}^{2d} \) is the concatenation of forward and backward hidden states at time step \( j \). We use the final hidden state as a representation of the patient's diagnostic history:
\begin{equation}
    z^{\text{codes}}_i = h_{i,m_i} \in \mathbb{R}^{2d}
\end{equation}

This vector is then projected to a lower-dimensional embedding \( {z'_i}^{\text{codes}} \in \mathbb{R}^d \) through a linear layer \( W_{\text{proj}} \in \mathbb{R}^{d \times 2d} \) to match the lab representation dimension for fusion:

\begin{equation}
    z'_i{}^{\text{codes}} = W_{\text{proj}} z^{\text{codes}}_i
\end{equation}
The projected embedding \( z'^{\text{codes}}_i \in \mathbb{R}^d \) is used in the cross-modal fusion module to integrate diagnostic information with the patient’s lab history.

\subsubsection{Cross-Attention Fusion Module}
Simple multimodal fusion via concatenation fails to account for the structural and temporal differences between laboratory trajectories and diagnosis codes. To address this, we introduce a cross-attention fusion mechanism that explicitly models inter-modal dependencies and enables feature-level alignment across modalities.

Let \( z^{\text{labs}}, {z'}^{\text{codes}} \in \mathbb{R}^{d} \) denote the lab and code embeddings for a given patient. To capture bidirectional interactions, we compute two cross-attention maps using standard query–key–value projections. For codes-to-labs attention, we define:
\begin{align}
Q_{\text{codes}} &= {z'}^{\text{codes}} {W}_{Q}^{c}, \quad K_{\text{labs}} = z^{\text{labs}} W_{K}^{l}, \quad V_{\text{labs}} = z^{\text{labs}} W_{V}^{l}\nonumber \\
A_{\text{codes} \rightarrow \text{labs}} &= \text{softmax} \left( \frac{Q_{\text{codes}} K_{\text{labs}}^\top}{\sqrt{d}} \right)
\end{align}
Similarly, labs-to-codes attention is computed as:
\begin{align}
Q_{\text{labs}} &= z^{\text{labs}} W_{Q}^{l}, \quad K_{\text{codes}} = {z'}^{\text{codes}} W_{K}^{c}, \quad V_{\text{codes}} = {z'}^{\text{codes}} W_{V}^{c} \nonumber \\
A_{\text{labs} \rightarrow \text{codes}} &= \text{softmax} \left( \frac{Q_{\text{labs}} K_{\text{codes}}^\top}{\sqrt{d}} \right)
\end{align}
where, \( W_{Q}^{l}, W_{K}^{l}, W_{V}^{l}, W_{Q}^{c}, W_{K}^{c}, W_{V}^{c} \in \mathbb{R}^{d \times d} \) are learnable projection matrices. 
 
These attention maps allow each modality to capture the most relevant features from the other modality. Specifically, the lab embedding is updated using lab-to-code attention, and the code embedding is updated using code-to-lab attention.
\begin{align}
\tilde{z'}^{\text{codes}} &= A_{\text{labs} \rightarrow \text{codes}} \cdot V_{\text{codes}} \\
\tilde{{z}}^{\text{labs}} &= A_{\text{codes} \rightarrow \text{labs}} \cdot V_{\text{labs}}
\end{align}
 The resulting vectors \( \tilde{z}^{\text{labs}} \) and \( \tilde{{z'}}^{\text{codes}} \) capture the most informative features from the other modality. Finally, the fused patient representation is obtained by concatenating the two enhanced embeddings. 

\begin{equation}
    z^{\text{fused}} = [\tilde{z}^{\text{labs}} \, \| \, \tilde{z'}^{\text{codes}}]
\end{equation}

This fused embedding is subsequently passed to multi-layer perceptron (MLP) for downstream tasks.

\section{Experimental Setup}

\subsection{Datasets} 
We collected a dataset of 9,950 patients from a private hospital, including 2,015 patients diagnosed with PDAC and 7,935 control patients. The dataset spans up to 14 years of patient trajectories and includes demographics, laboratory results, and diagnosis codes. The cancer cohort was identified using ICD-10 codes. The ICD system is structured hierarchically, ranging from broad categories such as \texttt{C}: ``Neoplasms'' to more specific disease codes like \texttt{C25.0} ``Malignant neoplasm of head of pancreas''. We considered patients with at least one diagnosis code under \texttt{C25} category (excluding \texttt{C25.4}: ``Malignant neoplasm of endocrine pancreas'') as PDAC cases. For the control group, we sampled patients from the hospital's EHR system, matching them to PDAC patients by sex and age at diagnosis while ensuring that they had no history of any \texttt{C25} diagnosis. Our analyses focused on longitudinal laboratory measurements and the chronological sequence of clinical events represented by diagnosis codes.  All diagnosis-code-based analyses were conducted using the first three characters of the ICD codes to reduce sparse representations of diagnoses and focus on broader categories.

\subsection{Tasks and Data Processing}

In this work, we develop models for the early detection of PDAC at three different prediction windows: 6 months, 9 months, and 12 months prior to diagnosis. We focus on the 6-month detection task as our primary evaluation and use the 9- and 12-month windows to further assess the efficacy of our multimodal pipeline. 

\subsubsection{Data Processing:}
We considered 35 routinely collected laboratory tests from four routinely collected panels, along with diagnosis codes. For each prediction task, we excluded all patient data occurring within the task-specific window (e.g., 6, 9, or 12 months) prior to the PDAC diagnosis date for cancer cases and prior to the most recent clinical encounter for matched controls.   To ensure path trajectory construction for the NCDE models, we removed patients with fewer than two measurements of the same laboratory test across their medical history. For diagnosis codes, we filtered out rare ICD-10 codes that appeared in fewer than 5\% of patients.  After preprocessing, the final cohort for the 6-month PDAC detection task included $4,694$ patients,  comprising $535$ PDAC cases and $4,159$ matched controls. The cohorts for the 9-month and 12-month prediction tasks were slightly smaller due to the stricter exclusion windows, resulting in $3,612$ patients ($418$ PDAC cases and $3,194$ controls) for the 9-month prediction task and $3,470$ patients ($372$ PDAC cases and $3,098$ controls) for the 12-month prediction task.

\subsection{Implementation Details and Baselines}
The implementation and training details of our method are availabe in Appendix  \ref{append:implement}. We also compare our method to three competitive baselines for early PDAC detection, which we reproduce using their publicly available source code:
\paragraph{CancerRiskNet} (\citealp{placido2023deep}): CancerRiskNet is a transformer-based architecture trained on long disease trajectories derived from diagnosis codes. It employs a modified positional encoding scheme that computes the position of each disease code token based on the time interval between the diagnosis date and the date of pancreatic cancer diagnosis.
\paragraph{Grouped Neural Network (GrpNN)} (\citealp{park2022deep}): GrpNN is a grouped neural network architecture designed to handle irregular, multivariate laboratory time series data. It independently embeds each variable's time-series into a lower-dimensional space. A Poisson-based random masking strategy is applied to the input during training to model the effects of information missingness.
\paragraph{Composite Neural Network (CompositeNN)} (\citealp{park2023structured}): CompositeNN is a structured deep embedding framework that groups laboratory features into organ-specific composites. Each group is processed through a dedicated neural network, and the resulting embeddings are concatenated to form the final representation used for early PDAC prediction.

\subsection{Evaluation Metrics}
To evaluate predictive performance for early PDAC detection, we use the Area Under the Curve (AUC) and the Area Under the Precision-Recall Curve (AUPRC) as metrics.  To assess model interpretability, we leverage Integrated Gradients (IG) ~\citep{sundararajan2017axiomatic} to analyze the contribution of individual diagnosis codes in identifying PDAC cases. Additionally, we study the attention weights from the Self-Attention layer in the labs processing module to evaluate the relative importance of different laboratory panels on the predictions. 

\section{Results} 
In this section, we present a comprehensive evaluation of our approach against the proposed baselines for early detection of PDAC at 6 months in advance (Section~\ref{subsec:early-PDAC}) and at 9 and 12 months in advance (Section~\ref{subsec:longer-horizon}). We further include an ablation study to assess the contribution of each component of our approach (Section~\ref{subsec:ablation-study}). Lastly, we analyze the interpretability of our model (Section~\ref{subsec:intrep-analysis}) by identifying key diagnosis-code-based and laboratory-based features associated with early PDAC detection.

\begin{table}[t]
\centering
\caption{Predictive performance of our proposed method and the baselines for 6-month early detection of PDAC. We specify the modalities used by each method: $\mathcal{L}$ refers to laboratory measurements, and $\mathcal{C}$ refers to diagnosis codes. Performance is summarized using AUC and AUPRC, reported as mean $\pm$ standard deviation across five-fold cross-validation. We also report DeLong's test p-value to compare our model's AUC performance to the baselines. }
\label{tab:predictive-perf-6M-PDAC}
\setlength{\tabcolsep}{10pt} 
\begin{tabular}{ccccc c}
\toprule
\textbf{Model} & \multicolumn{2}{c}{\textbf{Modality}} & \textbf{AUC} & \textbf{AUPRC} & \textbf{p-value} \\
\midrule
 & \(\mathcal{L}\) & \(\mathcal{C}\) & & & \\
\midrule
CancerRiskNet  &  & \checkmark & 0.657 $\pm$ 0.025 & 0.205 $\pm$ 0.032 & $4.97 \times 10^{-9}$ \\
CompositeNN    & \checkmark &  & 0.538 $\pm$ 0.025 & 0.145 $\pm$ 0.013 & $6.65 \times 10^{-39}$ \\
GrpNN          & \checkmark &  & 0.664 $\pm$ 0.026 & 0.265 $\pm$ 0.051 & $2.97 \times 10^{-27}$ \\
\midrule
\textbf{Ours}  & \checkmark & \checkmark & \textbf{0.738 $\pm$ 0.012} & \textbf{0.272 $\pm$ 0.011} & --- \\
\bottomrule
\end{tabular}
\end{table}

\subsection{Early Detection of PDAC at 6-month prior to actual clinical diagnosis}
\label{subsec:early-PDAC}
In Table \ref{tab:predictive-perf-6M-PDAC}, we compare the predictive performance of our proposed approach to several strong baselines for early detection of PDAC. The evaluation is based on AUC and AUPRC metrics. We note first that our approach is the only that jointly leveraging both modalities. Our method achieves significant improvements in predictive performance, with a 12.33\% increase in AUC over CancerRiskNet, the best baseline utilizing diagnosis codes ($\mathcal{C}$), and an 11.14\% increase over GrpNN, the best baseline using laboratory measurements ($\mathcal{L}$). Further, our approach shows the smallest standard deviation in performance across the five folds for both metrics. This indicates more consistent performance. These results highlight the advantages of jointly learning both modalities for the early detection PDAC.
\begin{table}[t]
\centering
\caption{Predictive performance of our proposed method and the baselines for early detection of PDAC. AUC is reported at 6, 9 and 12 months prior to diagnosis, shown as mean $\pm$ standard deviation across five-fold cross-validation.}
\label{tab:predictive-perf-PDAC-timepoints}
\setlength{\tabcolsep}{14pt} 
\begin{tabular}{lccc}
\toprule
\textbf{Model} & \textbf{AUC (6M)} & \textbf{AUC (9M)} & \textbf{AUC (12M)} \\
\midrule
\textbf{Cohort Size} & 4694 Patients & 3612 Patients & 3470 Patients\\
\midrule
LR  & 0.645 $\pm$ 0.015& 0.625 $\pm$ 0.044 & 0.622 $\pm$ 0.020 \\
CancerRiskNet  & 0.657 $\pm$ 0.025& 0.646 $\pm$ 0.036 & 0.589 $\pm$ 0.042 \\
GrpNN          & 0.664 $\pm$ 0.026& 0.648 $\pm$ 0.019 & 0.645 $\pm$ 0.040 \\
\midrule
\textbf{Ours}  & \textbf{0.738 $\pm$ 0.012} & \textbf{0.686 $\pm$ 0.021} & \textbf{0.680 $\pm$ 0.017} \\
\bottomrule
\end{tabular}
\end{table}

\subsection{Early Detection of PDAC at 9-month and 12-month in advanace}
\label{subsec:longer-horizon}

We next evaluate the ability of our method to detect PDAC at longer horizons (9 and 12 months prior to diagnosis) and compare its performance against two competitive benchmarks and Logistic Regression (LR). Table~\ref{tab:predictive-perf-PDAC-timepoints} summarizes AUC scores across the three timepoints. As the prediction window increases, the cohort size decreases due to fewer patients having sufficient historical data (4694 at 6 months, 3612 at 9 months, and 3470 at 12 months). Despite this reduction in cohort size, our model consistently outperforms the three baselines. At 9 months prior to diagnosis, our method achieves AUC improvements of 6.2\% and 5.9\% over CancerRiskNet and GrpNN, respectively. At 12 months in advance, it maintains strong performance, with AUC gains of 15.5\% over CancerRiskNet and 5.4\% over GrpNN. These results further highlight that the multimodal approach learns better representation even when data becomes more limited.
\begin{table}[h]
\centering
\caption{\textbf{Ablation Analysis.} We evaluate the impact of different model components (labs module, codes module, and fusion mechanism) on predictive performance in the six-month early detection task. Additionally, we examine the effect of modeling patient behavior through observational intensity variables and using BioGPT embeddings.}
\label{tab:abaltion-components}
\setlength{\tabcolsep}{15pt}
\begin{tabular}{lcc}
\toprule
\textbf{Modules or Constraints} & \textbf{AUC} & \textbf{AUPRC} \\
\midrule
Labs Module Only  & 0.673 $\pm$ 0.037 & 0.222 $\pm$ 0.042 \\
Codes Module Only   & 0.695 $\pm$ 0.022 & 0.251 $\pm$ 0.029 \\
w/o Cross-Attention       & 0.715 $\pm$ 0.014 & 0.256 $\pm$ 0.011 \\
w/o BioGPT        & 0.705 $\pm$ 0.035 & 0.242 $\pm$ 0.020 \\
w/o Observational Intensity           & 0.726 $\pm$ 0.018 & 0.269 $\pm$ 0.015 \\
\bottomrule                                                     
\textbf{Ours (All Components)} & \textbf{0.738 $\pm$ 0.012} & \textbf{0.272 $\pm$ 0.011} \\
\bottomrule
\end{tabular}
\end{table}

\begin{table}[]
\centering
\caption{\textbf{Fusion Method Comparison.} We compare different fusion strategies for combining lab and code modalities in the six-month early detection task.}
\label{tab:fusion-methods}
\setlength{\tabcolsep}{15pt}
\begin{tabular}{lcc}
\toprule
\textbf{Fusion Method} & \textbf{AUC} & \textbf{AUPRC} \\
\midrule
Concatenation & 0.7200~$\pm$~0.0137 & 0.2559~$\pm$~0.0096 \\
Concatenation + Self-Attention & 0.7283~$\pm$~0.0143 & 0.2650~$\pm$~0.0161 \\
Bilinear Fusion & 0.7274~$\pm$~0.0139 & \textbf{0.2731~$\pm$~0.0083} \\
\textbf{Cross-Attention (Ours)} & \textbf{0.7378~$\pm$~0.0121} & 0.2723~$\pm$~0.0111 \\
\bottomrule
\end{tabular}
\end{table}

\subsection{Ablation Study}
\label{subsec:ablation-study}
Table \ref{tab:abaltion-components} summarizes the results of our ablation analysis. We begin by evaluating the impact of individual model components, including training the labs-only module and the diagnosis-codes-only module, removing the cross-attention mechanism, and omitting the observational intensity features. First, Our labs-only module outperforms GrpNN by 1.36\% in AUC, demonstrating that modeling irregular laboratory time series with NCDEs effectively captures informative patterns, surpassing the Poisson distribution masking approach in GrpNN to handle missing data. Second, we observe a substantial improvement of 5.78\% in AUC from training our codes-only module compared to CancerRiskNet. This improvement indicates that compressed BioGPT embeddings of ICD-10 codes provide strong representations of diagnosis codes. Third, combining the two modules through a simple concatenation  between laboratory data and diagnosis-code feature vectors leads to significant performance improvements of 6.24\% over the labs-only module and 2.88\% over the codes-only module. By introducing the cross-attention mechanism, we observe a boost in AUC of 3.22\% above the model that concatenates the vector representations of the two modalities. This suggests that learning cross-modal interactions between the modalities is important. When BioGPT embeddings are excluded, the model learns diagnosis code embeddings from random initialization, resulting in a 4.45\% drop in AUC. This highlights the benefits of transferring knowledge from large pretrained language models. Incorporating observational intensity results in a  slight improvement of 1.65\% in AUC. This shows that adding these features helps the model learn better patient behavior patterns beyond what is captured by the shared representation. Finally, we observe a notable reduction in standard deviation across the five evaluation folds. This indicates that the two modalities provide complementary information, allowing the model to achieve stable performance.  

In addition to the component-wise ablation, we conducted a comparative analysis of alternative fusion strategies for combining laboratory and diagnosis-code modalities (Table~\ref{tab:fusion-methods}). All fusion methods achieved strong performance (AUC~$\geq$~0.72). Simple concatenation of feature vectors resulted in the lowest performance among the tested strategies. Adding a self-attention layer on top of the concatenated vectors provided a moderate improvement, suggesting that even modeling intra-modal interactions can enhance the joint representation. Bilinear fusion performed competitively, achieving the highest AUPRC (0.2731), but slightly underperformed in AUC compared to our method. Our cross-attention fusion approach achieved the best overall performance with an AUC of 0.7378 and a strong AUPRC of 0.2723.

\begin{table}[t]
\centering
\renewcommand{\arraystretch}{1.2} 
\caption{Top ICD-10 codes ranked by Integrated Gradients (IG) attribution values on the test dataset across five data split folds. Positive IG values correspond to strong predictors of PDAC in the early detection task.}
\label{tab:icd10_ig_table}
\begin{tabular}{lp{11cm}r}
\hline
\textbf{Code} & \textbf{Description} & \textbf{IG} \\
\hline
K86 & Chronic Pancreatitis & 0.359 \\
\hline
K85 & Acute pancreatitis & 0.319 \\
\hline
J20 & Acute bronchitis & 0.301 \\
\hline
L57 & Skin changes due to chronic exposure to ultraviolet radiation & 0.300 \\
\hline
R89 & Abnormal findings in specimens from other organs, systems and tissues & 0.291 \\
\hline
L82 & Seborrhoeic keratosis & 0.271 \\
\hline
L73 & Follicular disorders & 0.271 \\
\hline
G57 & Mononeuropathies of the lower limb & 0.268 \\
\hline
K63 & Other diseases of intestine & 0.262 \\
\hline
R32 & Unspecified urinary incontinence & 0.259 \\
\hline
\end{tabular}
\end{table}

\subsection{Interpretability Analysis}
\label{subsec:intrep-analysis}
To assess the quality of our model's predictions and the relevance of the learned representations to PDAC, we compute IG attributions for all unique diagnosis codes observed in at least 10 patients from the test set across a five-fold data split. Table \ref{tab:icd10_ig_table} lists the top 10 ICD-10 codes ranked by their IG attribution scores. The two highest-ranked codes, K86 (``Chronic Pancreatitis") and K85 (``Acute pancreatitis"), are clinically well-established risk factors for PDAC (\citealp{gandhi2022chronic}), as chronic and recurrent pancreatitis significantly increase the likelihood of developing pancreatic cancer. Interestingly, the third-ranked code is J20 (``Acute bronchitis"). While this code is not a known risk factor for PDAC, \citealp{huang2023chronic}  reported a correlation between bronchitis and worse PDAC progression. Several skin-related codes also appear among the top contributors, including L57 (``Skin changes due to chronic exposure to ultravioletradiation"), L82 (``Seborrhoeic keratosis"), and L73 (``Other follicular disorders"). Although these are not directly linked to PDAC, shared genetic factors such as CDKN2A mutations have been associated with both PDAC (\citealp{kimura2021role}) and certain skin conditions (\citealp{fargnoli2010mc1r}), suggesting a possible biological connection. L57 also indicates radiation exposure, pointing to an underlying serious illness. We also observe that K63 (``Other diseases of the intestine") appears top-ranked codes, which aligns with PDAC representation by non-specific gastrointestinal symptoms (\citealp{lan2023challenges}). Finally, R89 (“Abnormal findings in specimens from other organs, systems and tissues”) ranks fifth. This may reflect early, unexplained abnormalities documented by clinicians before a formal PDAC diagnosis is made.

\begin{figure}[t]
    \centering
    \includegraphics[width=0.8\textwidth]{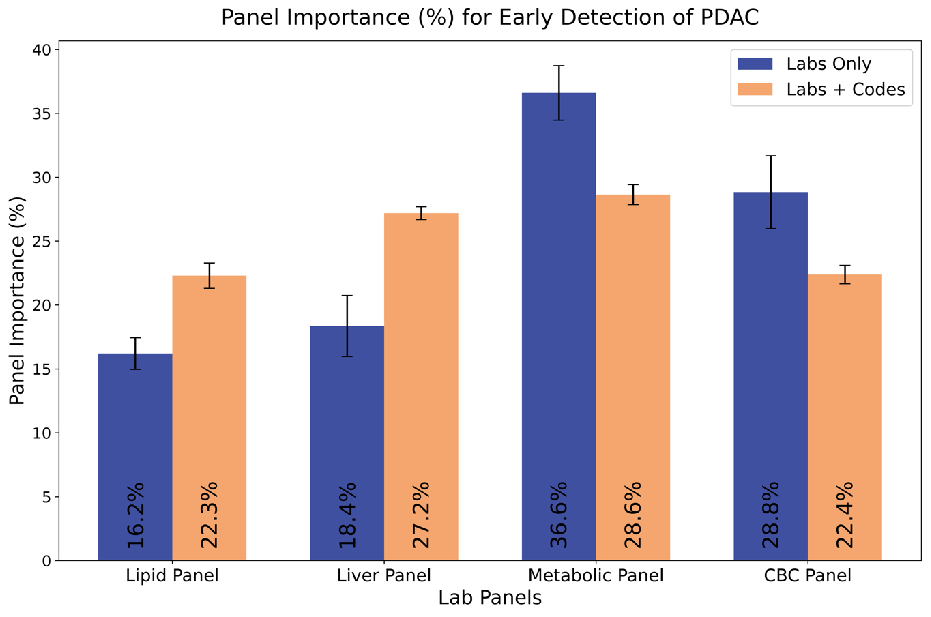}
    \caption{Importance of each panel in the predictive model for our proposed architecture and the labs only module}
    \label{fig:panel_importance}
\end{figure}

We also examine which laboratory panels the model relies on most when making predictions by analyzing the self-attention weights. Specifically, we compare the panel-wise importance when the model is trained using only laboratory data versus when it is trained using both laboratory data and diagnosis code trajectories. Figure~\ref{fig:panel_importance} summarizes the panel attention weights across five-fold cross-validation. When trained on laboratory data only, the model places the most emphasis on the metabolic panel (36.6\%) and the CBC panel (28.8\%), with lower importance given to the liver and lipid panels. In contrast, when trained with both modalities, the model distributes attention more evenly, with greater emphasis on the liver panel (27.7\%) and the metabolic panel (28.6\%). This shift is desirable, as both panels are known to reflect early PDAC signals. Liver panel abnormalities may indicate biliary obstruction from tumors in the pancreatic head, while metabolic disruptions such as  glucose dysregulation are recognized early markers of disease onset.

\section{Discussion} 


We presented a novel approach for the early detection of PDAC using routinely collected laboratory measurements and diagnosis code histories. We proposed a multimodal formulation that captrues the interplay between laboratory data and diagnosis codes to learn more informative representations of disease history. Additionally, we introduced a new module to encode longitudinal laboratory data as irregular time series at the panel level by grouping labs into functional panels, learning panel-specific representation through NCDEs and inter-panel representations by a self-attention mechanism. We also introduced a module to learn rich feature representations from the temporal trajectories of diagnosis codes by generating compressed BioGPT embeddings of the codes and learning their sequence representation with a bidirectional recurrent network.

The findings show that our framework significantly improves  early detection of PDAC at different prediction times up to one year in advance. Our method consistently outperforms several competitive baselines across five-fold cross-validation. The ablation study further highlights the contribution of each component in our architecture. First, modeling laboratory time series using continuous-time dynamics, along with panel-level and patient-level representations, outperforms approaches that impose priors on missingness or rely on fixed-length sequences. Second, incorporating pretrained language models with recurrent networks yields better diagnosis trajectory representations than training from scratch. Third, learning rich cross-modal representations between labs and diagnosis codes further boosts predictive performance.

From an interpretability standpoint, our analysis showed that the model's behavior aligns well with clinical knowledge. Specifically, the top contributing diagnosis codes for early detection include ``Acute Pancreatitis" and ``Chronic Pancreatitis", both are clinically recognized risk factors for PDAC. Other codes, such as R32 (``Unspecified urinary incontinence"), are indirectly linked to PDAC through associations with aging and diabetes-related complications. The model also highlights additional codes that could serve as potential biomarkers for identifying high-risk patients, including J20 ("Acute bronchitis"), R89 ("Abnormal findings in specimens from other organs, systems, and tissues"), L57 (``exposure to UV radiation"), and other skin-related codes (L82, L73). Although these codes are not directly linked to PDAC, existing clinical studies have reported associations between their underlying conditions and specific aspects of pancreatic cancer. Finally,  our analysis revealed that combining the two modalities during training enables the model to assign greater importance to panels more closely associated with PDAC, such as the liver and metabolic panels.  We hope these insights encourage further focused clinical research to explore these potential links.

\paragraph{Limitations and Future Work}

Our method has several limitations that will inform the direction of future work. First, the cross-attention fusion mechanism is currently applied at the aggregated representation level of labs and diagnosis code trajectories. However, different lab panels may have distinct relationships with specific diagnosis codes, and capturing these fine-grained interactions could improve performance and interpretability. Second, in building our diagnosis code processing module, we leverage BioGPT to generate feature embeddings. While BioGPT can produce clinically meaningful representations, it is not explicitly trained on sequences of structured diagnosis codes, which may limit its ability to capture disease trajectory representations fully. Third, our approach has been trained and evaluated on a single dataset, which may limit the generalizability of our findings. Pancreatic cancer is a rare disease, and publicly available datasets with sufficient temporal resolution and diagnostic detail are scarce, which constrained our ability to perform external validation.

In future work, we plan to address these limitations. We will extend our model to better capture fine-grained interactions between specific lab panels and related diagnosis codes. We also aim to develop a dedicated approach for learning representations of disease trajectories that takes the temporal and hierarchical properties of diagnosis sequences into account. Finally, we are actively collaborating with multiple hospitals to acquire more diverse datasets and evaluate our models across different clinical settings, with the goal of improving robustness and generalizability.

\section{Conclusion}

Our work demonstrates that routinely collected laboratory measurements and diagnosis codes can be effectively used for the early detection of PDAC, enabling scalable risk screening across a broader population. Our model substantially improves predictive accuracy by learning a cross-modal representation for sequences of diagnosis codes and longitudinal lab data. The model builds on clinically established risk factors to generate the predictions while also uncovering novel diagnostic signals that may serve as potential early biomarkers. Furthermore, it leverages relevant lab panels to capture physiologically meaningful patterns, enhancing predictive performance.

\acks{This work was supported by the Jump ARCHES grant P349. We thank Marlene Robles Granda, David McGrew, and Roopa Foulger for their assistance in data collection. We also thank Chang Hu and Krishnakant Saboo for their valuable discussions and helpful comments on the paper.
}

\bibliography{sample}

\begin{thebibliography}{35}
\providecommand{\natexlab}[1]{#1}
\providecommand{\url}[1]{\texttt{#1}}
\expandafter\ifx\csname urlstyle\endcsname\relax
  \providecommand{\doi}[1]{doi: #1}\else
  \providecommand{\doi}{doi: \begingroup \urlstyle{rm}\Url}\fi

\bibitem[Appelbaum et~al.(2021)Appelbaum, Cambronero, Stevens, Horng, Pollick, Silva, Haneuse, Piatkowski, Benhaga, Duey, et~al.]{appelbaum2021development}
Limor Appelbaum, Jos{\'e}~P Cambronero, Jennifer~P Stevens, Steven Horng, Karla Pollick, George Silva, Sebastien Haneuse, Gail Piatkowski, Nordine Benhaga, Stacey Duey, et~al.
\newblock Development and validation of a pancreatic cancer risk model for the general population using electronic health records: An observational study.
\newblock \emph{European Journal of Cancer}, 143:\penalty0 19--30, 2021.

\bibitem[Ax(2021)]{axPlatform2021}
Ax.
\newblock {Ax · Adaptive Experimentation Platform}, 2021.
\newblock URL \url{https://ax.dev/}.
\newblock Accessed [Date you accessed the website].

\bibitem[Blackford et~al.(2024)Blackford, Canto, Dbouk, Hruban, Katona, Chak, Brand, Syngal, Farrell, Kastrinos, et~al.]{blackford2024pancreatic}
Amanda~L Blackford, Marcia~Irene Canto, Mohamad Dbouk, Ralph~H Hruban, Bryson~W Katona, Amitabh Chak, Randall~E Brand, Sapna Syngal, James Farrell, Fay Kastrinos, et~al.
\newblock Pancreatic cancer surveillance and survival of high-risk individuals.
\newblock \emph{JAMA oncology}, 10\penalty0 (8):\penalty0 1087--1096, 2024.

\bibitem[Che et~al.(2018)Che, Purushotham, Cho, Sontag, and Liu]{che2018recurrent}
Zhengping Che, Sanjay Purushotham, Kyunghyun Cho, David Sontag, and Yan Liu.
\newblock Recurrent neural networks for multivariate time series with missing values.
\newblock \emph{Scientific reports}, 8\penalty0 (1):\penalty0 6085, 2018.

\bibitem[Chen et~al.(2021)Chen, Cherry, Nalawade, Qiao, Kumar, Lowy, Simpson, and Murphy]{chen2021clinical}
Qinyu Chen, Daniel~R Cherry, Vinit Nalawade, Edmund~M Qiao, Abhishek Kumar, Andrew~M Lowy, Daniel~R Simpson, and James~D Murphy.
\newblock Clinical data prediction model to identify patients with early-stage pancreatic cancer.
\newblock \emph{JCO clinical cancer informatics}, 5:\penalty0 279--287, 2021.

\bibitem[Chen et~al.(2018)Chen, Rubanova, Bettencourt, and Duvenaud]{chen2018neural}
Ricky~TQ Chen, Yulia Rubanova, Jesse Bettencourt, and David~K Duvenaud.
\newblock Neural ordinary differential equations.
\newblock \emph{Advances in neural information processing systems}, 31, 2018.

\bibitem[Chen et~al.(2023)Chen, Phuc, Nguyen, Burton, Lin, Lin, Lu, Hsu, Cheng, and Hsu]{chen2023novel}
Shih-Min Chen, Phan~Thanh Phuc, Phung-Anh Nguyen, Whitney Burton, Shwu-Jiuan Lin, Weei-Chin Lin, Christine~Y Lu, Min-Huei Hsu, Chi-Tsun Cheng, and Jason~C Hsu.
\newblock A novel prediction model of the risk of pancreatic cancer among diabetes patients using multiple clinical data and machine learning.
\newblock \emph{Cancer Medicine}, 12\penalty0 (19):\penalty0 19987--19999, 2023.

\bibitem[Cichosz et~al.(2024)Cichosz, Jensen, Hejlesen, Henriksen, Drewes, and Olesen]{cichosz2024prediction}
Simon~Lebech Cichosz, Morten~Hasselstr{\o}m Jensen, Ole Hejlesen, Stine~Dam Henriksen, Asbj{\o}rn~Mohr Drewes, and S{\o}ren~Schou Olesen.
\newblock Prediction of pancreatic cancer risk in patients with new-onset diabetes using a machine learning approach based on routine biochemical parameters.
\newblock \emph{Computer methods and programs in biomedicine}, 244:\penalty0 107965, 2024.

\bibitem[Espona-Fiedler et~al.(2024)Espona-Fiedler, Patthey, Lindblad, Sarr{\'o}, and {\"O}hlund]{espona2024overcoming}
Margarita Espona-Fiedler, Cedric Patthey, Stina Lindblad, Irina Sarr{\'o}, and Daniel {\"O}hlund.
\newblock Overcoming therapy resistance in pancreatic cancer: New insights and future directions.
\newblock \emph{Biochemical Pharmacology}, page 116492, 2024.

\bibitem[Fargnoli et~al.(2010)Fargnoli, Gandini, Peris, Maisonneuve, and Raimondi]{fargnoli2010mc1r}
Maria~Concetta Fargnoli, Sara Gandini, Ketty Peris, Patrick Maisonneuve, and Sara Raimondi.
\newblock Mc1r variants increase melanoma risk in families with cdkn2a mutations: a meta-analysis.
\newblock \emph{European journal of cancer}, 46\penalty0 (8):\penalty0 1413--1420, 2010.

\bibitem[Gandhi et~al.(2022)Gandhi, De~La~Fuente, Murad, and Majumder]{gandhi2022chronic}
Sonal Gandhi, Jaime De~La~Fuente, Mohammad~Hassan Murad, and Shounak Majumder.
\newblock Chronic pancreatitis is a risk factor for pancreatic cancer, and incidence increases with duration of disease: a systematic review and meta-analysis.
\newblock \emph{Clinical and translational gastroenterology}, 13\penalty0 (3):\penalty0 e00463, 2022.

\bibitem[Haab et~al.(2024)Haab, Qian, Staal, Jain, Fahrmann, Worthington, Prosser, Velokokhatnaya, Lopez, Tang, et~al.]{haab2024rigorous}
Brian Haab, Lu~Qian, Ben Staal, Maneesh Jain, Johannes Fahrmann, Christine Worthington, Denise Prosser, Liudmila Velokokhatnaya, Camden Lopez, Runlong Tang, et~al.
\newblock A rigorous multi-laboratory study of known pdac biomarkers identifies increased sensitivity and specificity over ca19-9 alone.
\newblock \emph{Cancer letters}, 604:\penalty0 217245, 2024.

\bibitem[Horn et~al.(2020)Horn, Moor, Bock, Rieck, and Borgwardt]{horn2020set}
Max Horn, Michael Moor, Christian Bock, Bastian Rieck, and Karsten Borgwardt.
\newblock Set functions for time series.
\newblock In \emph{International Conference on Machine Learning}, pages 4353--4363. PMLR, 2020.

\bibitem[Huang et~al.(2023)Huang, Hammelef, Sabitsky, Ream, Khalilieh, Zohar, Lavu, Bowne, Yeo, and Nevler]{huang2023chronic}
Rachel Huang, Emma Hammelef, Matthew Sabitsky, Carolyn Ream, Saed Khalilieh, Nitzan Zohar, Harish Lavu, Wilbur~B Bowne, Charles~J Yeo, and Avinoam Nevler.
\newblock Chronic obstructive pulmonary disease is associated with worse oncologic outcomes in early-stage resected pancreatic and periampullary cancers.
\newblock \emph{Biomedicines}, 11\penalty0 (6):\penalty0 1684, 2023.

\bibitem[Huang et~al.(2015)Huang, Xu, and Yu]{huang2015bidirectional}
Zhiheng Huang, Wei Xu, and Kai Yu.
\newblock Bidirectional lstm-crf models for sequence tagging.
\newblock \emph{arXiv preprint arXiv:1508.01991}, 2015.

\bibitem[Janssens and Wasser(2013)]{janssens2013managing}
Pim~MW Janssens and Gerd Wasser.
\newblock Managing laboratory test ordering through test frequency filtering.
\newblock \emph{Clinical Chemistry and Laboratory Medicine (CCLM)}, 51\penalty0 (6):\penalty0 1207--1215, 2013.

\bibitem[Jia et~al.(2023)Jia, Kundrot, Palchuk, Warnick, Haapala, Kaplan, Rinard, and Appelbaum]{jia2023pancreatic}
Kai Jia, Steven Kundrot, Matvey~B Palchuk, Jeff Warnick, Kathryn Haapala, Irving~D Kaplan, Martin Rinard, and Limor Appelbaum.
\newblock A pancreatic cancer risk prediction model (prism) developed and validated on large-scale us clinical data.
\newblock \emph{EBioMedicine}, 98, 2023.

\bibitem[Kane et~al.(2023)Kane, King, Esserman, Latham, Greene, and Ganz]{kane2023compressed}
Michael~J Kane, Casey King, Denise Esserman, Nancy~K Latham, Erich~J Greene, and David~A Ganz.
\newblock A compressed large language model embedding dataset of icd 10 cm descriptions.
\newblock \emph{BMC bioinformatics}, 24\penalty0 (1):\penalty0 482, 2023.

\bibitem[Kidger et~al.(2020)Kidger, Morrill, Foster, and Lyons]{kidger2020neural}
Patrick Kidger, James Morrill, James Foster, and Terry Lyons.
\newblock Neural controlled differential equations for irregular time series.
\newblock \emph{Advances in neural information processing systems}, 33:\penalty0 6696--6707, 2020.

\bibitem[Kimura et~al.(2021)Kimura, Klein, Hruban, and Roberts]{kimura2021role}
Hirokazu Kimura, Alison~P Klein, Ralph~H Hruban, and Nicholas~J Roberts.
\newblock The role of inherited pathogenic cdkn2a variants in susceptibility to pancreatic cancer.
\newblock \emph{Pancreas}, 50\penalty0 (8):\penalty0 1123--1130, 2021.

\bibitem[Lan et~al.(2023)Lan, Robin, Kasnik, Wong, and Abdel-Rahman]{lan2023challenges}
Xiaoyang Lan, Gabrielle Robin, Jessica Kasnik, Grace Wong, and Omar Abdel-Rahman.
\newblock Challenges in diagnosis and treatment of pancreatic exocrine insufficiency among patients with pancreatic ductal adenocarcinoma.
\newblock \emph{Cancers}, 15\penalty0 (4):\penalty0 1331, 2023.

\bibitem[Leiphrakpam et~al.(2025)Leiphrakpam, Chowdhury, Zhang, Bajaj, Dhir, and Are]{leiphrakpam2025trends}
Premila~Devi Leiphrakpam, Sanjib Chowdhury, Michelle Zhang, Varnica Bajaj, Mashaal Dhir, and Chandrakanth Are.
\newblock Trends in the global incidence of pancreatic cancer and a brief review of its histologic and molecular subtypes.
\newblock \emph{Journal of Gastrointestinal Cancer}, 56\penalty0 (1):\penalty0 71, 2025.

\bibitem[Loshchilov and Hutter(2017)]{loshchilov2017decoupled}
Ilya Loshchilov and Frank Hutter.
\newblock Decoupled weight decay regularization.
\newblock \emph{arXiv preprint arXiv:1711.05101}, 2017.

\bibitem[Luo et~al.(2022)Luo, Sun, Xia, Qin, Zhang, Poon, and Liu]{luo2022biogpt}
Renqian Luo, Liai Sun, Yingce Xia, Tao Qin, Sheng Zhang, Hoifung Poon, and Tie-Yan Liu.
\newblock Biogpt: generative pre-trained transformer for biomedical text generation and mining.
\newblock \emph{Briefings in bioinformatics}, 23\penalty0 (6):\penalty0 bbac409, 2022.

\bibitem[Morrill et~al.(2021)Morrill, Kidger, Yang, and Lyons]{morrill2021neural}
James Morrill, Patrick Kidger, Lingyi Yang, and Terry Lyons.
\newblock Neural controlled differential equations for online prediction tasks.
\newblock \emph{arXiv preprint arXiv:2106.11028}, 2021.

\bibitem[Park et~al.(2022)Park, Artin, Lee, Pumpalova, Ingram, May, Park, Hur, and Tatonetti]{park2022deep}
Jiheum Park, Michael~G Artin, Kate~E Lee, Yoanna~S Pumpalova, Myles~A Ingram, Benjamin~L May, Michael Park, Chin Hur, and Nicholas~P Tatonetti.
\newblock Deep learning on time series laboratory test results from electronic health records for early detection of pancreatic cancer.
\newblock \emph{Journal of biomedical informatics}, 131:\penalty0 104095, 2022.

\bibitem[Park et~al.(2023)Park, Artin, Lee, May, Park, Hur, and Tatonetti]{park2023structured}
Jiheum Park, Michael~G Artin, Kate~E Lee, Benjamin~L May, Michael Park, Chin Hur, and Nicholas~P Tatonetti.
\newblock Structured deep embedding model to generate composite clinical indices from electronic health records for early detection of pancreatic cancer.
\newblock \emph{Patterns}, 4\penalty0 (1), 2023.

\bibitem[Placido et~al.(2023)Placido, Yuan, Hjaltelin, Zheng, Haue, Chmura, Yuan, Kim, Umeton, Antell, et~al.]{placido2023deep}
Davide Placido, Bo~Yuan, Jessica~X Hjaltelin, Chunlei Zheng, Amalie~D Haue, Piotr~J Chmura, Chen Yuan, Jihye Kim, Renato Umeton, Gregory Antell, et~al.
\newblock A deep learning algorithm to predict risk of pancreatic cancer from disease trajectories.
\newblock \emph{Nature medicine}, 29\penalty0 (5):\penalty0 1113--1122, 2023.

\bibitem[Rubanova et~al.(2019)Rubanova, Chen, and Duvenaud]{rubanova2019latent}
Yulia Rubanova, Ricky~TQ Chen, and David~K Duvenaud.
\newblock Latent ordinary differential equations for irregularly-sampled time series.
\newblock \emph{Advances in neural information processing systems}, 32, 2019.

\bibitem[Sharma et~al.(2018)Sharma, Smyrk, Levy, Topazian, and Chari]{sharma2018fasting}
Ayush Sharma, Thomas~C Smyrk, Michael~J Levy, Mark~A Topazian, and Suresh~T Chari.
\newblock Fasting blood glucose levels provide estimate of duration and progression of pancreatic cancer before diagnosis.
\newblock \emph{Gastroenterology}, 155\penalty0 (2):\penalty0 490--500, 2018.

\bibitem[Shukla and Marlin(2021)]{shukla2021multi}
Satya~Narayan Shukla and Benjamin~M Marlin.
\newblock Multi-time attention networks for irregularly sampled time series.
\newblock \emph{arXiv preprint arXiv:2101.10318}, 2021.

\bibitem[Singhi et~al.(2019)Singhi, Koay, Chari, and Maitra]{singhi2019early}
Aatur~D Singhi, Eugene~J Koay, Suresh~T Chari, and Anirban Maitra.
\newblock Early detection of pancreatic cancer: opportunities and challenges.
\newblock \emph{Gastroenterology}, 156\penalty0 (7):\penalty0 2024--2040, 2019.

\bibitem[Sundararajan et~al.(2017)Sundararajan, Taly, and Yan]{sundararajan2017axiomatic}
Mukund Sundararajan, Ankur Taly, and Qiqi Yan.
\newblock Axiomatic attribution for deep networks.
\newblock In \emph{International conference on machine learning}, pages 3319--3328. PMLR, 2017.

\bibitem[{World Health Organization}(2019)]{WHO2019ICD10}
{World Health Organization}.
\newblock \emph{International Statistical Classification of Diseases and Related Health Problems (10th Revision)}.
\newblock World Health Organization, 2019.
\newblock URL \url{https://icd.who.int/}.

\bibitem[Zhang et~al.(2021)Zhang, Zeman, Tsiligkaridis, and Zitnik]{zhang2021graph}
Xiang Zhang, Marko Zeman, Theodoros Tsiligkaridis, and Marinka Zitnik.
\newblock Graph-guided network for irregularly sampled multivariate time series.
\newblock \emph{arXiv preprint arXiv:2110.05357}, 2021.

\end{thebibliography}

\newpage
\appendix
\section{Appendix A. Implementation and Training Details}
\label{append:implement}

  \paragraph{Code}   The code to reproduce our experiments is available at \url{https://github.com/MosbahAouad/EarlyPDAC-MML}.
 
  \paragraph{Data Splits} For all tasks, models were evaluated using five-fold cross-validation. The data was split into 64\% for training, 16\% for validation, and 20\% for testing in each fold.

  \paragraph{Data Normalization}  All laboratory measurements were normalized per channel before being input into the NCDE. Each channel was standardized to have zero mean and unit variance.

  \paragraph{Optimizer} In all experiments, we used the AdamW optimizer \citep{loshchilov2017decoupled} with a learning rate of 0.0005 and a batch size of 512. We applied $\ell_2$ regularization with a weight of $w = 0.0005$ and used a linear learning rate warm-up over the first 5 epochs. If the validation loss did not decrease for 7 consecutive epochs, the learning rate was halved. Training was terminated if the validation loss failed to improve for 21 consecutive epochs, and the model checkpoint with the lowest validation loss was retained.

  \paragraph{ODE Solver}  NCDE requires an ODE solver to model the hidden state dynamics. We used the Dormand–Prince 5 method (\texttt{"dopri5"}) to evaluate the integral in the NCDE formulation. The absolute and relative tolerances were set to $10^{-4}$ and $10^{-2}$, respectively.
  
  \paragraph{Model Architecture} Our proposed architecture consists of three main components: a labs-processing module, a codes-processing module, and a cross-attention module. The detailed model hyperparameters are provided in Table~\ref{tab:proposed-model-architect}.
  
  \paragraph{Hyperparameter Tuning} We used Bayesian Optimization to tune the hyperparameters of each component in our proposed approach. Specifically, we used the Adaptive Experimentation Platform \citep{axPlatform2021} with 10 optimization trials.

For the labs-processing module (NCDE), the search space included: \texttt{hidden\_dimension} in $[32, 256]$, \texttt{hidden\_hidden\_dimension} in $[32, 256]$, \texttt{num\_layers} in $\{1, 2, 3, 4\}$, \texttt{dropout} in $[0, 3]$, and \texttt{self\_attention\_num\_heads} in $\{2, 4, 8\}$.

For the codes-processing module (Bi-LSTM), the search space included: \texttt{BioGPT\_embedding\_dim} in $\{50, 100, 1000\}$, \texttt{bi\_lstm\_layer\_num} in $\{1, 2, 3, 4\}$, \texttt{bi\_lstm\_layer\_hidden} in $[64, 256]$, and \texttt{dropout} in $[0, 30]$.

\begin{table}[t]
\centering
\caption{Model Configuration Parameters}
\label{tab:proposed-model-architect}
\small 
\begin{tabular}{@{}c|c|c@{}}
\toprule
\textbf{Component} & \textbf{Parameter}                              & \textbf{Value}              \\ \midrule
\multirow{10}{*}{Labs Processing Module (NCDE)} 
                   & interpolation\_method                           & Natural Cubic Splines       \\
                   & input\_time\_series\_length                     & 100                         \\
                   & hidden\_dimension                               & 128                         \\
                   & hidden\_hidden\_dimension                       & 128                         \\
                   & num\_layers                                     & 1                           \\
                   & dropout                                         & 0.1                         \\
                   & self\_attention\_value                          & 128                         \\
                   & self\_attention\_query                          & 128                         \\
                   & self\_attention\_key                            & 128                         \\
                   & self\_attention\_heads                          & 2                           \\ \midrule
\multirow{4}{*}{Codes Processing Module (Bi-LSTM)} 
                   & BioGPT\_embedding\_dim                          & 50                          \\
                   & bi\_lstm\_layer\_num                            & 1                           \\
                   & bi\_lstm\_layer\_hidden                         & 64                          \\
                   & dropout                                         & 0                           \\ \midrule
\multirow{7}{*}{Cross-Attention Module} 
                   & cross\_attention\_labs\_to\_codes\_value        & 128                         \\
                   & cross\_attention\_labs\_to\_codes\_key          & 128                         \\
                   & cross\_attention\_labs\_to\_codes\_query        & 128                         \\
                   & cross\_attention\_codes\_to\_labs\_value        & 128                         \\
                   & cross\_attention\_codes\_to\_labs\_key          & 128                         \\
                   & cross\_attention\_codes\_to\_labs\_query        & 128                         \\
                   & mlp\_classification\_layer\_dim                 & 256                         \\
\bottomrule
\end{tabular}
\end{table}

\section*{Appendix B. Labs Data Description}
In this work, we develop our methods on dataset of routinely collected longitudinal lab measurements. We use 35 labs split into four main panels: Metabolic panel (10 labs), CBC panel (15 labs), lipid panel (5 labs), and liver panel (5 labs). The detailed labs of each panel are listed below: 

\begin{itemize}
    \item \textbf{Lipid Panel}: Cholesterol, LDL, HDL, Triglycerides, Total Protein
    \item \textbf{Liver Panel}: ALP, ALT, AST, Albumin, Total Bilirubin
    \item \textbf{Metabolic Panel}: Glucose, Anion, BUN, CO\textsubscript{2}, Calcium, Chloride, Creatinine, Potassium, Sodium, A/G Ratio
    \item \textbf{CBC}: Absolute Basophils, Absolute Eosinophils, Absolute Lymphocytes, Absolute Monocytes, Absolute Neutrophils, MCHC, MCH, MCV, MPV, WBC, RBC, RDW, Platelet, Hemoglobin, Hematocrit
\end{itemize}

\section*{Appendix C. Additional Analysis}
\subsection{Subgroup Analysis by Gender}
We conducted a subgroup analysis by gender and report AUC for males and females. Our method achieved comparable performance across both groups, with the differences diminishing as the prediction window increased:

\begin{itemize}
  \item \textbf{6-month window:} \\
  Females (n~$\approx$~519): AUC = 0.7449~$\pm$~0.0223 \\
  Males (n~$\approx$~420): AUC = 0.7298~$\pm$~0.0074
  \vspace{0.5em}
  
  \item \textbf{9-month window:} \\
  Females (n~$\approx$~394): AUC = 0.6893~$\pm$~0.0331 \\
  Males (n~$\approx$~325): AUC = 0.6816~$\pm$~0.0227
  \vspace{0.5em}
  
  \item \textbf{12-month window:} \\
  Females (n~$\approx$~394): AUC = 0.6781~$\pm$~0.0184 \\
  Males (n~$\approx$~323): AUC = 0.6756~$\pm$~0.0447
\end{itemize}

\subsection{Additional Metrics}
We provide AUPRC results across all three prediction time points in the table below. Our proposed approach consistently outperforms baselines at each prediction point.
\begin{table}[h]
\centering
\caption{AUPRC of our proposed method and the baselines for early detection of PDAC. AUPRC is reported at 6, 9 and 12 months prior to diagnosis, shown as mean $\pm$ standard deviation across five-fold cross-validation.}
\label{tab:auprc-perf-PDAC-timepoints}
\setlength{\tabcolsep}{14pt} 
\begin{tabular}{lccc}
\toprule
\textbf{Model} & \textbf{AUPRC (6M)} & \textbf{AUPRC (9M)} & \textbf{AUPRC (12M)} \\
\midrule
CancerRiskNet  & 0.2053 $\pm$ 0.0324 & 0.2072 $\pm$ 0.0261 & 0.1661 $\pm$ 0.0271 \\
GrpNN          & 0.2650 $\pm$ 0.0510 & 0.2290 $\pm$ 0.0403 & 0.2068 $\pm$ 0.0563 \\
\midrule
\textbf{Ours}  & \textbf{0.2720 $\pm$ 0.0110} & \textbf{0.2335 $\pm$ 0.0298} & \textbf{0.2142 $\pm$ 0.0312} \\
\bottomrule
\end{tabular}
\end{table}
\end{document}